# A context-based geoprocessing framework for optimizing meetup location of multiple moving objects along road networks

Shaohua Wang[1,2], Song Gao[3], Xin Feng[2], Alan T. Murray[2], Yuan Zeng[4]

1. Institute of Geographic Sciences and Natural Resources Research, Chinese Academy of Science, Beijing, China

2. Department of Geography, University of California, Santa Barbara, CA, USA

3. Department of Geography, University of Wisconsin, Madison, WI, USA

4. Esri Inc., 380 New York Street, Redlands, CA, USA

[1]Abstract: Given different types of constraints on human life, people must make decisions that satisfy social activity needs. Minimizing costs (i.e., distance, time, or money) associated with travel plays an important role in perceived and realized social quality of life. Identifying optimal interaction locations on road networks when there are multiple moving objects (MMO) with space-time constraints remains a challenge. In this research, we formalize the problem of finding dynamic ideal interaction locations for MMO as a spatial optimization model and introduce a context-based geoprocessing heuristic framework to address this problem. As a proof of concept, a case study involving identification of a meetup location for multiple people under traffic conditions is used to validate the proposed geoprocessing framework. Five heuristic methods with regard to efficient shortest-path search space have been tested. We find that the R* tree-based algorithm performs the best with high quality solutions and low computation time. This framework is implemented in a GIS environment to facilitate integration with external geographic contextual information, e.g., temporary road barriers, points of interest (POI), and real-time traffic information, when dynamically searching for ideal meetup sites. The proposed method can be applied in trip planning, carpooling services, collaborative interaction, and logistics management.

Keywords: route planning; network analysis; spatial optimization; shortest-path heuristics.

---

## 1. Introduction

Human movements and interactions in space and time lay a foundation for supporting our economic and social activities. The increasing development of location-based information communication technologies and services leads to an upsurge of geogrxaphic information; humans can use this information to support decision making on movements and interactions in space and time. Given different types of pressures and commitments in human life, especially space-time constraints (Hägerstraand 1970, Schwanen and Kwan 2008), people must make decisions to satisfy social activity participation needs and travel behavior. By integrating time geography theory and spatial analysis methods, researchers have made advances and achievements in evaluating spatial accessibility, human activities and interactions in both physical space and virtual space with spatiotemporal constraints (Kim and Kwan 2003; Miller 1991, 2005a; Shaw and Yu 2009). Recently, the economic, environmental, and social needs of urban sustainable transportation and sustainable cities have facilitated the development of a ridesharing economy (Black et al. 2002, Kennedy et al. 2005). There exist a variety of ridesharing scenarios under trip-based or activity-based considerations (Wang et al. 2016). High occupancy rates per vehicle through ridesharing can help reduce traffic. However, the search for potential interaction locations (e.g., intermediate meetups or pick-ups) involving multiple moving objects or vehicle trajectories with space-time constraints remains a challenge. For example, how could friends driving from different work locations find an intermediate meetup location for ride-sharing to a party or an event with limited parking capacity, or for handing over keys or important documents before heading to different destinations? In addition, contextual information, such as traffic congestion, temporal barriers, points of interest (POI), weather and other environmental factors, play an important role in enabling and limiting movements (Buchin et al. 2012, Demšar et al., 2015, Dodge et al. 2016, Siła-Nowicka et al. 2016).

In this research, we aim to formalize the problem of finding an optimal meetup location for multiple moving objects (MMO) and introduce a context-based geoprocessing framework to solve this as a spatial optimization problem with efficient path-finding heuristics. The contributions of our research are three-fold. First, we formalize and solve the MMO dynamic meetup location problem mathematically with regard to different search scenarios (section 3). Second, we propose a context-based geoprocessing framework for optimizing the MMO meetup location problem with road network constraints. This proposed framework computes an ideal meetup location by taking into account the spatiotemporal context (e.g., traffic delays with different road types) from nearly real-time data streams or other geospatial datasets (e.g., POI database) and GIS services. Third, the computational needs involved in our approach are minimal. The proposed heuristics are found to identify high quality solutions and are more efficient than classic spatial optimization solution techniques.

The remainder of paper is organized as follows. First, in section 2, we present the literature review on spatiotemporal movement analysis and spatial optimization approaches. In section 3, we formalize the MMO dynamic meetup location problem, present optimization approaches, and introduce a context-based geoprocessing framework with heuristic techniques for finding the dynamic meetup location on road networks for MMO. In section 4, we explain our computational procedures as well as the data used for the testing. In section 5, we test our framework on three case studies for different MMO on road networks with traffic and POI information and demonstrate the effectiveness and computational efficiency of our proposed geoprocessing framework. Finally, we conclude this work with some consideration on the potential of this work for trip planning and other routing services, and present our vision for future research in section 6.

## 2. Related work

Finding a good meetup location for a group of people is prevalent in daily life whether it is a business meeting, carpooling, socialization, etc. Identifying an ideal meetup location is an important research problem in transportation and spatial query systems (Kuijpers & Othman 2009; Yan et al. 2011, 2015). Travel costs can be measured by distance, time, and/or price. In terms of distance and time, which are our focus in this study, there are generally three main approaches: *Manhattan*, *Euclidean* and *Networks* metrics.

In Euclidean space, the optimal meeting point problem can be cast as a classic Weber problem (Cooper 1968, Drezner and Goldman 1991). It is one of the most fundamental and well-known problems in location theory. The goal is to find the location that minimizes the sum of the weighted transportation costs. Direct numerical solutions have been developed (Weiszfeld 1937, Tellier 1972). The work by Vardi and Zhang (2001) showed that a modified version of Weiszfeld's algorithm for solving the Weber problem converged monotonically to a unique solution. A gradient-descent framework was developed for finding an optimal location (Yan et al. 2011, 2015).

The Multi-Weber problem is an extension of the Weber problem involving multi-facilities (Plastria 1995). The multi-Weber problem concerns siting multiple facilities simultaneously to serve all demands as efficiently as possible (Cooper, 1968). The facilities and demands in the Weber problem can be structured to correspond to meetup locations and objects' origins and destinations, respectively.

Kuijpers & Othman (2009) studied space-time prisms for modelling uncertainty in moving objects along a road network and developed an algorithm to compute and visualize the 3D space-time prisms constrained by network travel. They also presented a solution to the *alibi* query, which asks whether two moving objects can potentially meet. Further, if they can, the meetup location is found by intersecting two network-time prisms. Since the space-time prisms capture all possible locations of a moving object or person between two anchor points (i.e., the location of home and work) (Hägerstraand 1970, Couclelis et al. 1987, Miller 2005b) as well as the time schedule given the maximum travel speed, one has to assume that the locations and times of two anchor points are fixed. This is a limitation for classical space-time prisms. In practice, people's travel schedules should have a certain degree of inherent flexibility for departures and arrivals. To address this issue, Kuijpers et al. (2010) developed the concept of "anchor regions". The concepts of network-time-prisms are widely used in recent sustainable mobility studies for urban public transportation planning (Song et al. 2017). In addition, Yan et al. (2015) designed a base-line algorithm and two-phase convex-hull pruning techniques for solving the meeting point problem for a multi-point set. However, all aforementioned methods don't necessarily grantee an optimal solution.

From an optimization perspective, Hakimi (1965) showed that there exists a subset containing $p$ vertices on the network containing an optimal solution to the $p$-median problem. Therefore, the vertices of the network can be taken as a set of candidates. Xu & Jacobsen (2010) developed a base-line algorithm for finding an optimal meeting point for $Q$ pairs of origins and destinations on a road network $G=(V,E)$ with $V$ vertices and $E$ edges in query datasets. Yan et al. (2015) proved that the size of the search space for optimal meeting point query is $|V|+|Q|$. On road networks, we focus on how to reduce search space (i.e., scanned road vertices and segments) to find an optimal meeting location, exploring a number of different optimization heuristics (see section 3.4).

One key step in the meetup point query for a group of people is to find their shortest/fastest paths to a destination. Dijkstra's algorithm is frequently used to find the shortest path on road networks (Dijkstra 1959). Later, Hart et al. (1968) developed another efficient search strategy called A* by using heuristics to guide the search for minimum cost paths. In previous empirical studies along real-world road networks, A* has been found to outperform other approaches (Zeng & Church 2009). The efficiency of the original shortest path algorithms can be improved by further reducing the search space (Wagner & Willhalm 2007, Huang et al. 2007). Another study showed that the speed-up ratio of shortest-path computation is linearly related to a reduction in search space (Wang et al. 2013). Reducing the search space of vertices and road segments is a good strategy for finding an efficient solution to an optimal meeting point on road networks. Graph theory suggests potential search space reduction for finding the ideal meeting point based on bidirectional search (Goldberg & Harrelson 2005) and multi-level graph overlay (Holzer et al. 2009). Heuristic methods are often efficient for solving shortest-path problems (Huang et al. 2007; Zeng & Church 2009). In this study, we will present five heuristic methods (in section 3.4) for finding the ideal meetup point along road networks.

## 3. Methodology

In this section, we formally define the optimal meetup problem using Manhattan distance, formulate the intermediate meetup problem along road networks, and present a context-based geoprocessing framework using optimization heuristics to support location queries.

### *3.1 Problem definition using the Manhattan distance*

The simplest scenario is that multiple people need to determinate a location as their shared final destination to meet while minimizing the total travel cost (distance or time).

Notation:

*m: index of people;*

*$(X_m, Y_m)$: two-dimensional origin coordinates for person m;*

*(X, Y): two-dimensional coordinates for meetup location;*

*$W_m$: weight for travel cost of m;*

*$D_m$: distance from person m to the meetup location.*

The model can be mathematically expressed as follows:

Objective Function:  Minimize $\sum_m W_m D_m$ (1)

If we assume that the distance $D_m$ is a Manhattan distance, it can be expressed as:

$$D_m = |X - X_m| + |Y - Y_m| \quad (2)$$

The objective function becomes:

Minimize $\sum_m W_m (|X - X_m| + |Y - Y_m|)$ (3)

Assuming positive coordinate values, the objective function is equivalent to the following linear programming problem (Church & Murray 2009):

Minimize $\sum_m (W_m DX_m + W_m DY_m)$ (4)

Subject to:

$DX_m \geq X - X_m, \quad \forall m$ (5)

$DX_m \geq X_m - X, \quad \forall m$ (6)

$DY_m \geq Y - Y_m, \quad \forall m$ (7)

$DY_m \geq Y_m - Y, \quad \forall m$ (8)

$DX_m \geq 0, DY_m \geq 0, \quad \forall m$ (9)

$X \geq 0, Y \geq 0$ (10)

Objective (4) seeks the minimum total weighted distance. Constraints (5)-(8) define distance in both the *x* and *y* directions from each person to the selected meetup location. The distance variable $D_m$ is decomposed into two components, $DX_m$ for the *x* direction and $DY_m$ for the *y* direction. Constraints (9)-(10) are non-negativity requirements on decision variables. The Manhattan-distance-based formulation could be used for a comparison with the later introduced network-based formulation (section 3.2).

Spatial optimization problems are often solved by linear programming methods (Tong and Murray 2012), where a function of the variables is optimized subject to a set of equations that describe the constraints. In order to validate whether this model can be used to find an exact optimal meetup location, we tested it with zero tolerance for the optimality deviation using the GUROBI optimization solver[2]. For example, assuming four people from different origins want to meet at a location which minimizes the total travel distance, without road network information across a homogeneous space. Let the cost weight $W_m$ equal to 1 for all MMO. By applying the linear programming functions (4)~(10) using a group of MMO with integer coordinates, A (10,42), B (0,0), C (45, 33), D (5, 20), the optimal meetup location is (8,20). However, in real-world scenarios, we cannot simply solve this optimal location problem assuming a homogenous planar space. Geographic contexts including the land cover type (e.g., water bodies or land), the road network, transportation accessibility and other factors need to be considered.

---

[2] http://www.gurobi.com/

*3.2 Origin-Destination Intermediate Meetup on a Road Network*

Another type of meetup scenario is that the MMO are heading to different destinations after their intermediate meetup. For example, friends may want to pass keys or documents, drop-off kids, or grab coffee together on their way to the office. In such a case, people seek to maximize benefits while minimizing deviations from their initially planned routes in choosing an intermediate meetup location. However, deviations are part of the total travel costs. Given a road network, we can formalize this problem as follows:

Notation:

$i$ = *index of movement objects;*

$j$ = *index reference to a potential meetup location;*

$D_{i,j}$ = *shortest distance along the network from the origin of i to j;*

$D'_{j,i}$ = *shortest distance along the network from j to the destination of i;*

$W_{i,j}$ = *a travel cost weight for i moving from origin of i to meetup location j;*

$W'_{j,i}$ = *a travel cost weight for i moving from meetup location j to destination of i;*

$$Y_j = \begin{cases} 1, \text{if meetup selected at location } j; \\ 0, \text{otherwise.} \end{cases}$$

$$X_{i,j} = \begin{cases} 1, \text{if object } i \text{ goes to meetup location } j; \\ 0, \text{otherwise} \end{cases}$$

With this notation, a spatial optimization model structuring the meetup location on a road network is desired, minimizing the total travel costs (i.e., weighted network distance in our study) for all MMO. Note that the travel cost weight $W$ can take the contextual information such as traffic delay to update the weighted network cost between two nodes. The formulation is as follows:

$$\text{Minimize:} \quad \sum_i \sum_j (W_{i,j} * D_{i,j} + W'_{j,i} * D'_{j,i}) * X_{i,j} \quad (11)$$

$$\text{Subject to:} \quad \sum_j Y_j = 1 \quad (12)$$

$$X_{i,j} \leq Y_j, \quad \forall i, j \quad (13)$$

$$\sum_j X_{i,j} = 1, \quad \forall i \quad (14)$$

$$Y_i = \{0, 1\}, \quad \forall i \quad (15)$$

$$X_{i,j} = \{0, 1\}, \quad \forall i, j \quad (16)$$

Objective (11) seeks to minimize total weighted path distances including trips to and from a meetup location. Constraint (12) requires that exactly one meetup location on the road network is selected. Constraints (13) allow assignment only to the selected meetup location. Constraints (14) ensure that each person is assigned to only one meetup location. Constraints (15) and (16) impose binary restrictions on decision variables.

## *3.3 Geoprocessing Framework*

As discussed above, real-world scenarios suggest consideration of different geographic contexts which further increases computational complexity. A novel geoprocessing framework in GIS is introduced to solve this optimization problem. This provides flexibility for integrating external geospatial datasets and location-based services, but also eliminates the need for expensive commercial optimization solvers.

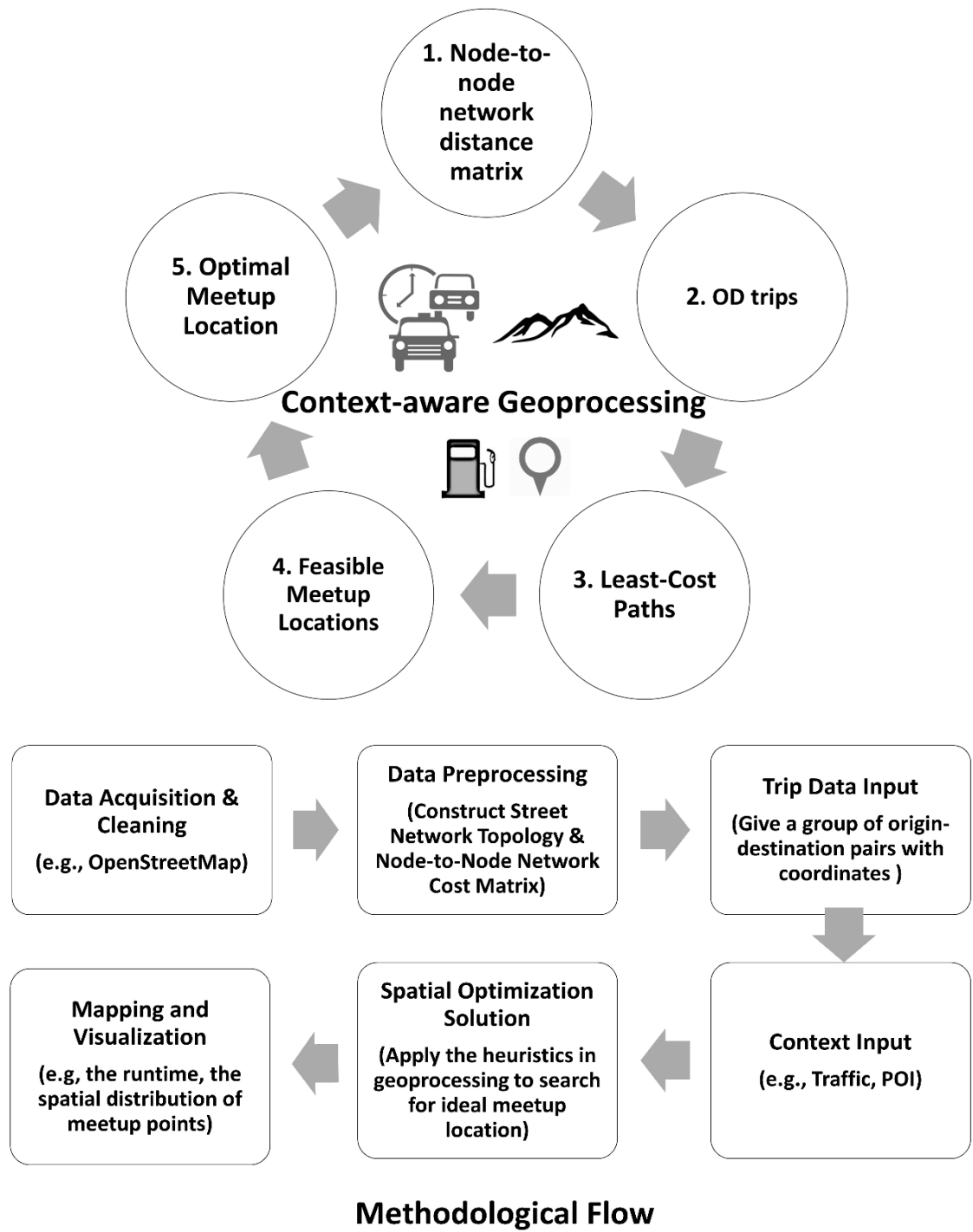

[Figure 1. A context-based geoprocessing framework with methodological workflow to find an ideal meetup location for multiple moving objects.]

As shown in Figure 1, firstly, pairwise intersection-to-intersection distance (or a travel cost matrix) on the road networks is needed, precomputed and stored on a computer or a processing server. Secondly, an input layer is planned origin-to-destination (OD) trips (with locations and time budgets) for multiple moving objects. Thirdly, the least-cost path (LCP) for every object can be computed based on either the shortest-path distance or minimum travel time. The classic Dijkstra (1959) or A* (Hart et al. 1968) algorithms can be used for finding shortest paths between nodes in a graph if road network information is available. Otherwise, visibility-graph-based approaches may be used for deriving shortest paths in continuous space (Hong and Murray 2013). If a meetup happens in the context of mountainous areas, accumulated cost surface with regard to both distance and slope, or viewshed analysis on digital elevation models, can be employed to compute the LCP (Douglas 1994; Stucky 1998). Fourthly, based on the search space containing the overlapping road intersections and segments or the corridors from the computation of LCP and spatial alternative routes (Lombard & Church 1993), we can find a feasible meetup zone that satisfies all interaction conditions for MMO. Finally, an optimal meetup location can be selected according to the objective function. In addition, the existence of additional contextual information, such as the availability of preferred POI categories for meeting, can affect the final decision of selecting a meetup location. Note that an "ideal" meetup location under different contexts may not necessarily be the "optimal" point that minimizes the objective travel cost. Such contextual information can be integrated in the geoprocessing framework. The required data input and processing methodologies are also presented in the flowchart (Figure 1). Examples will be further discussed in Section 5.

### *3.4 Heuristic optimization approaches on road networks*

Five heuristic methods have been developed and applied for finding an optimal meetup location for MMO along road networks. One key concept in our heuristic algorithms is the *search space*, which contains the unique identifiers (IDs) of visited road intersections and segments for the shortest-path between each pair of origin-to-destination. Search-space vectors are computed and indexed during the shortest-path search process. In this research, we implement the following five heuristic approaches and compare their performance for accuracy, computation time and expansion size of search space.

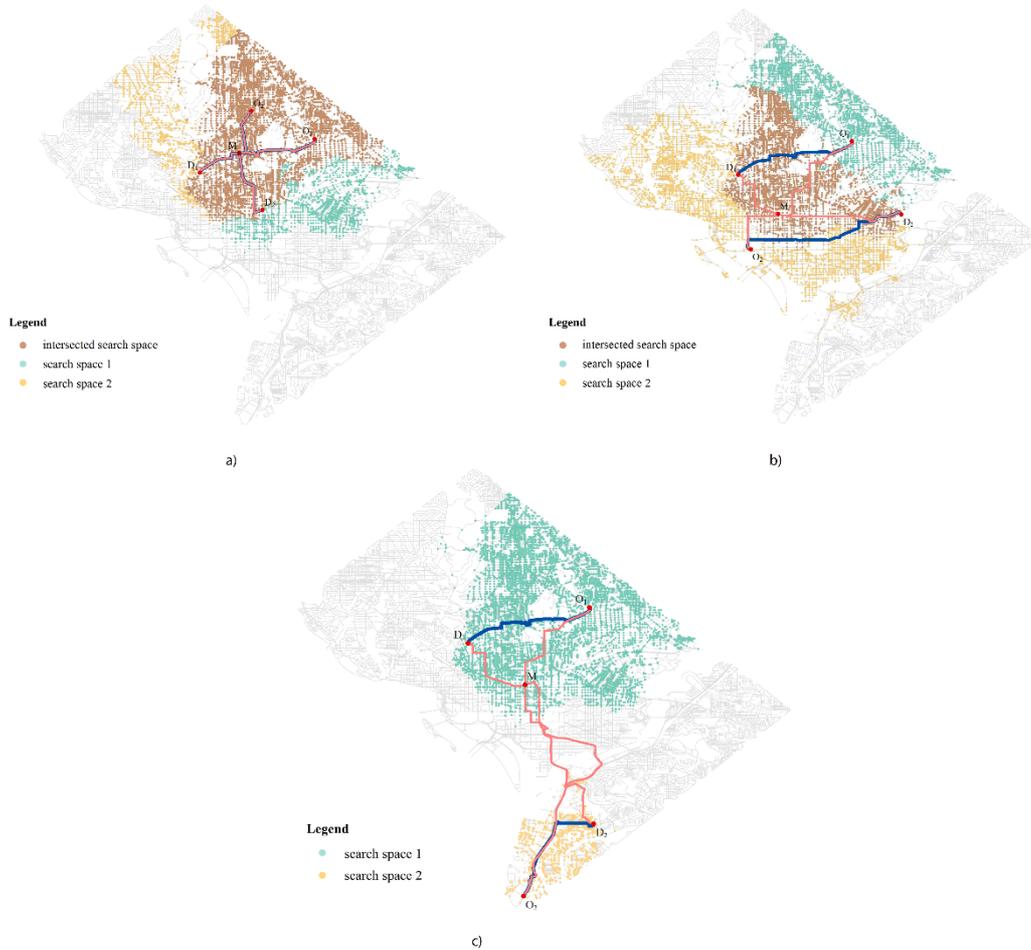

[Figure 2. Finding an optimal meetup location for two moving objects based on the shortest-path search space method with three scenarios: a] their shortest paths have shared segments and search spaces also overlap; b] only the shortest-path search spaces overlap; c] neither the shortest paths nor the search spaces overlap.]

- Shortest-path-search-space-based algorithm (SP): This was designed to reduce the search size of meeting point query. There are three scenarios using the SP method for two persons' meetup. First, the shortest paths for two persons travelling from their origins to destinations have shared road segments and their shortest-path search spaces also overlap (Figure 2a). Second, the shortest paths have no shared road segments but their shortest-path search spaces overlap (Figure 2b). Third, the shortest paths have no shared road segments and their shortest-path search spaces don't overlap either (Figure 2c). In the case of three MMO, as shown in Figure 3a, the meetup location can be found through a similar search manner. Using the SP method for a meetup location query, three steps are required to process: 1) Find the shortest path for each of all origin–destination pairs; 2) Find the intersections of their search spaces for getting common scanned nodes and segments on road networks; 3) Loop each node or segment in the overlapping search space to derive the meetup location by minimizing the sum of weighted distance or time cost for all MMO. If the search space apart, we will run the R* tree-based approach (see below) instead.

- Convex-hull-based algorithm (CH): This was designed for finding the minimum bounding geometry for a set of points in computational geometries (de Berg et al. 2000). In our research, as shown in Figure 3b, we apply the CH idea to reduce the search size for the meetup location query, although it cannot grantee optimum. Based on this algorithm, the key to determine the meetup point is to acquire the minimum convex hull of the point set which consists of all the trip origins and destinations (Yan et al. 2011). This algorithm saves search time by looping through each node in the convex hull, rather than the entire study area. Note that if the spatial distribution of origins and destinations span the entire study area, reducing the size of the search space may not be worth the effort.
- Diameter-point-based algorithm (DP): Another way to improve the efficiency of locating meetup point is to determine the search space by tracing the diameter points of the convex hull generated from a set of MMO trip origins and destinations (Aingworth et al. 1999). The diameter for a set of points is the greatest Euclidean distance between any two points in the set. With the two diameter points, the search space is constructed based on the union set of bidirectional shortest paths between them (in Figure 3c). Finally, an optimal meetup location is further selected from the intersections of road segments in the miniaturized search space by minimizing the total travel distance for all original MMO trips after they detour for an intermediate meetup.
- R* tree (RT): This has been formed to be efficient for organizing and indexing point data in computational geometries and GIS applications (Beckmann et al. 1990). In our experiments, an R* tree-based spatial indexing approach is used to bound a box query for locating meetup point efficiently (in Figure 3d). This box is the smallest bounding rectangle based on all origins and destinations.
- Euclidian-distance-based algorithm (ED): This is relied on finding a meetup point in Euclidean space to efficiently locate the meetup location along road networks. The k-dimensional tree is used for searching the k-nearest-neighbors (KNN) from the optimal meeting point by a Euclidian distance (Bentley 1975). One of those k-nearest vertices is defined as the ideal meetup point in the road network space. As shown in Figure 3e, the original meetup location based on the Euclidian distance is the orange point (M'), and then the ideal meetup location M along the road network can be further identified based on the KNN search. Note that *k* equals to one-tenth of all street nodes in this illustration case and the value may vary in different read-world transportation routing applications.

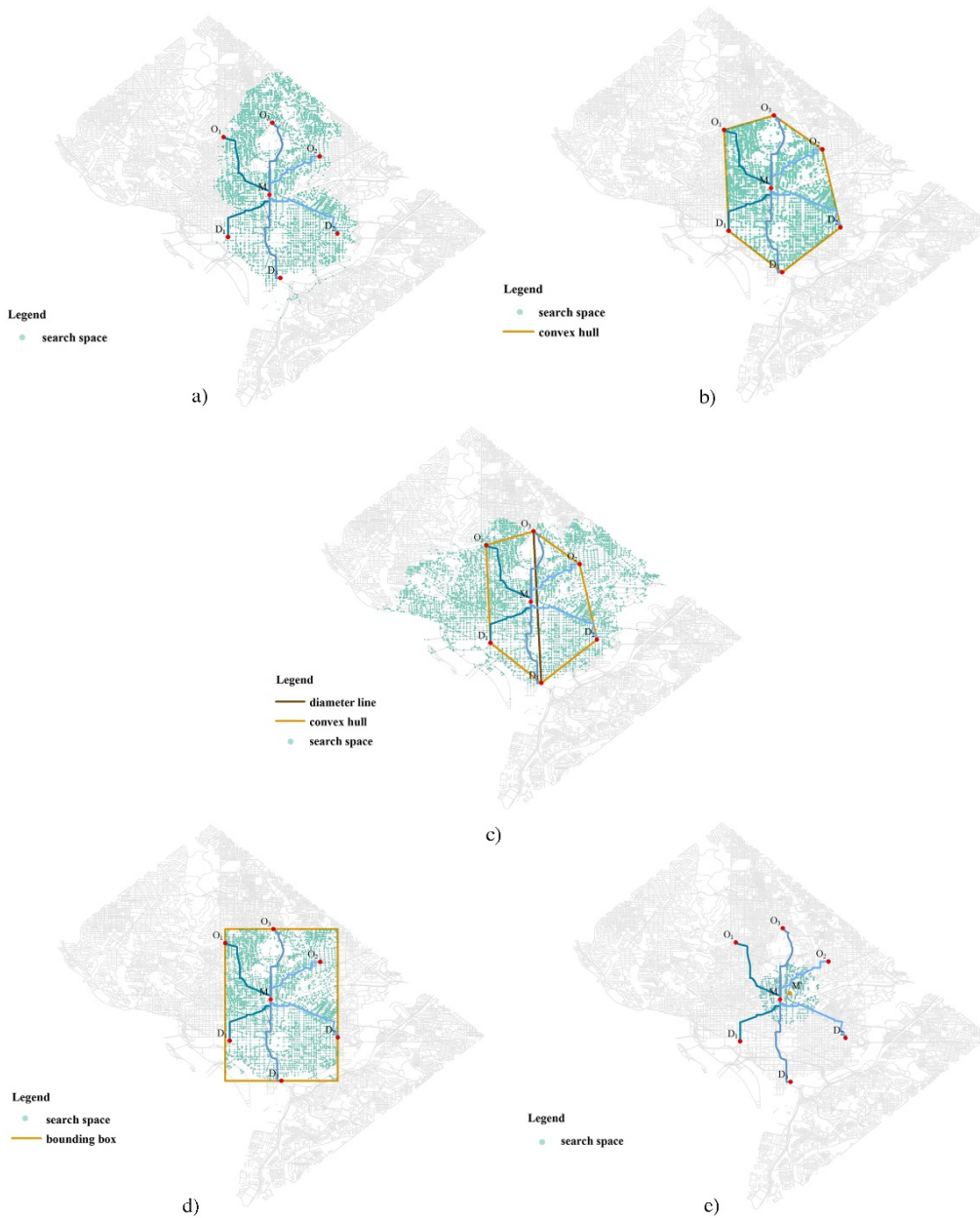

[Figure 3. Finding an optimal meetup location for three moving objects based on five different search space strategies: a) SP; b) CH; c) DP; d) RT; e) ED.]

## 4. Data and Computational Processing

Assessment of meetup locations for different scenarios on a road network for *Washington D.C.* serves as our study area[3]. This benchmark testing data contain 14,909 road segments and 9,559 nodes after data cleaning. Each road segment consists of information about road category, distance (a great circle distance), and estimated travel time. In addition, in order to add the traffic and POI context information, we also acquired OpenStreetMap street-segment data for the study area and constructed the node-edge topology network (with 18,363 nodes and 28,178 edges) from the raw trajectories and ways (Boeing 2017). The traffic simulation process will be further discussed in section 5.2. The limitations and potentials of using OpenStreetMap data in GIScience research and a comparison with authoritative road datasets have been discussed in the literature (Haklay 2010, Arsanjani et al. 2015).

As introduced in the geoprocessing framework section, the pairwise intersection-to-intersection distance and travel time over the road network must be precomputed and stored. Dijstra's algorithm is used along with a binary-heap data structure to compute the shortest-path distance costs. Network distance between each pair of nodes is stored in a node-cost-matrix. Shortest-path indices are used for constructing the road network distance matrix (Samet et al. 2008), but consumes very large CPU memory for large size road networks. An accelerated on-the-fly shortest-path computation strategy (Geisberger et al. 2008) is likely best for massive road networks (larger than one million nodes).

In order to study the overall optimality characteristics of "the sum of meetup cost based on road network distances", we derive meetup costs for all nodes on the road network and summarize the distribution surface. Figure 4 shows the meetup cost 3D surface using sums of network distances for all road intersections that serve as the meetup locations given two fixed pairs of origins and destinations (OD) for two moving objects used in Figure 6. The meetup locations are represented by *(X, Y)* coordinates, and the *Z* value refers to the sum of network distances for two objects to intermediately meet at the corresponding intersection *(X, Y)* on the road network. The bowl shape of the scatter surface shows the convexity property of the "meetup cost based on road network distances." We can use a fast-greedy algorithm for the optimal meetup location query on road networks by searching for the k-nearest-neighbors of road intersections during the geoprocessing.

---

[3] http://www.dis.uniroma1.it/challenge9/download.shtml

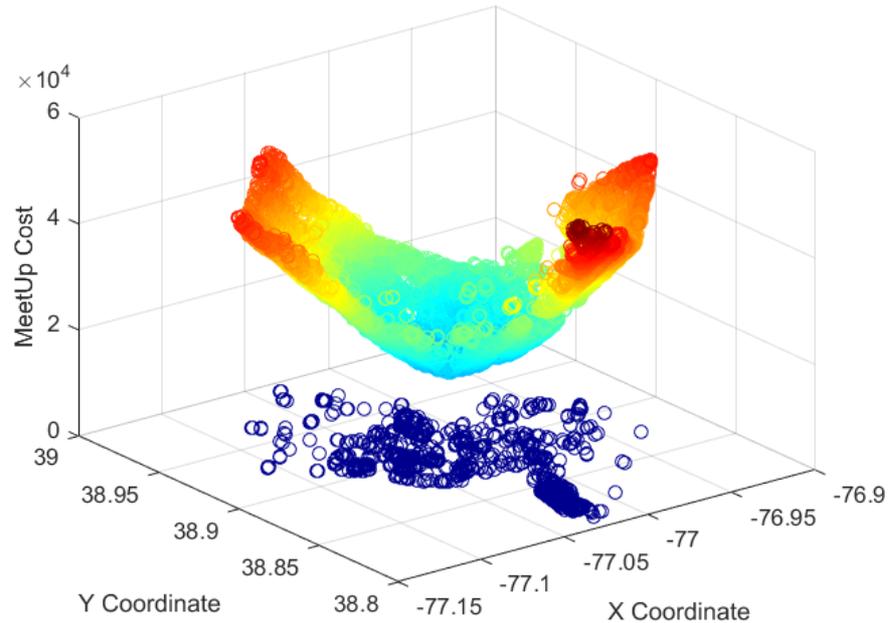

[Figure 4. The surface of the sums of road network distances for meetups located at the intersections on a road network given two fixed OD pairs.]

5. Analysis and Results

Our experiments were conducted on an Intel Core (TM) (3.40 GHz) computer running Windows 7 64-bit operating system with 8 GB of RAM. *SuperMap* and *ArcGIS* software were utilized for GIS data creation, management, manipulation, analysis and display. Esri's Arcpy, a Python geometry computation library, was relied upon for deriving spatial relationships and conditions needed in geoprocessing. Python code was written to use data inputs in structuring the spatial optimization model. GUROBI, an optimization solver, was employed to solve all problem instances with actual road network constraints for validation. The proposed heuristics have been integrated into our GIS processing framework for finding an optimal meetup location for MMO.

*5.1 Two pairs of MMO*

In the first group of experiments, we only consider the case of two moving objects meeting at one time. The aim is to find a meetup location at one intersection of the road segments. As mentioned in the methodology section, different search strategies are applied for three different shortest-path and search-space overlapping scenarios (see Figure 2). We randomly generated 1000 cases involving two pairs of origins and destinations for two moving objects. The spatial optimization model and five heuristic algorithms are applied for each case. The results of spatial optimization model can also be used for evaluating the correctness (accuracy) of the five heuristic algorithms since the spatial optimization model enables identification of a global optimal solution. If the objective value (i.e., the minimum total costs) derived from a heuristic geoprocessing approach is the same with that derived from the spatial optimization model, then it is successful.

Table 1. The results of finding optimal meetup cost objective value for 1000 simulated pairs of two moving objects using five different heuristic approaches: shortest-path-search-space-based (SP); convex-hull-based (CH); Diameter-point-based (DP); R* tree-based (RT); and Euclidian-distance-based (ED) algorithms.

| Methods | Number of found optimal cases | Number of missed cases | Accuracy |
|---|---|---|---|
| **SP** | 960 | 40 | 96.0 % |
| **CH** | 862 | 138 | 86.2 % |
| **DP** | 1000 | 0 | 100 % |
| **RT** | 996 | 4 | 99.6 % |
| **ED** | 962 | 38 | 96.2 % |

The results for the 1000 simulated cases are summarized in the Table 1. It shows that the DP heuristic algorithm has the highest accuracy of 100% since it identifies all 1000 optimal solutions. The RT, ED, SP heuristic algorithms also get very high accuracy of 99.6%, 96.2%, and 96.0%, but cannot find optimal meetup solutions for 4, 38, 40 cases out of total 1000 cases, respectively. The CH algorithm has the lowest accuracy, 86.2 %, since it cannot find optimal meetup solutions for 138 cases. Our speculation is that the shape and the size of the constructed convex-hull limits finding all possible overlapping parts of two shortest paths and the road intersections/segments search space for getting their shortest paths.

While ideally we want to get the optimal solution for each case in finding a meetup location for MMO in practical applications, it may be computationally expensive for deriving the best solution for a spatial optimization model. In our experiments, we need about 20 ~ 30 seconds to run the optimization package for finding an optimal solution with the precomputed intersection-to-intersection distance matrix of road networks; otherwise the runtimes take minutes or even longer without the physically stored distance matrix. However, with the support of different heuristic approaches, the computation efficiency improves significantly. As shown in Figure 5, the computation time for five aforementioned heuristic approaches is compared. Figure 5a shows the experimental runtime for each of the simulated 1000 cases and the variations in different cases clearly exist. Figure 5b displays the cumulative distribution function (CDF) for the runtime distributions. The RT and ED heuristic algorithms find the optimal meetup solution within about 1 millisecond for over 90% cases while the CH approach reaches the optimum over 85% cases within about 1 millisecond runtime. The SP approach runs slower than the RT, ED and CH approaches, taking over 15 milliseconds for 90% cases. The DP approach is the slowest, requiring over 30 milliseconds for 85% cases. Using a paired two-sample t-test, we found that the average time differences between all methods were statistically significant ($p<0.001$), except for between RT and ED which showed no significant difference ($p = 0.06$). In summary, the most accurate heuristic approach, DP, needs more run time than the other approaches. Decision making for an efficient optimal solution in practice might be a trade-off between computation time and accuracy, especially for large-scale road networks with millions of road intersections and segments; the RT and ED algorithms are reasonable alternatives that run much faster with higher accuracy compared to other approaches.

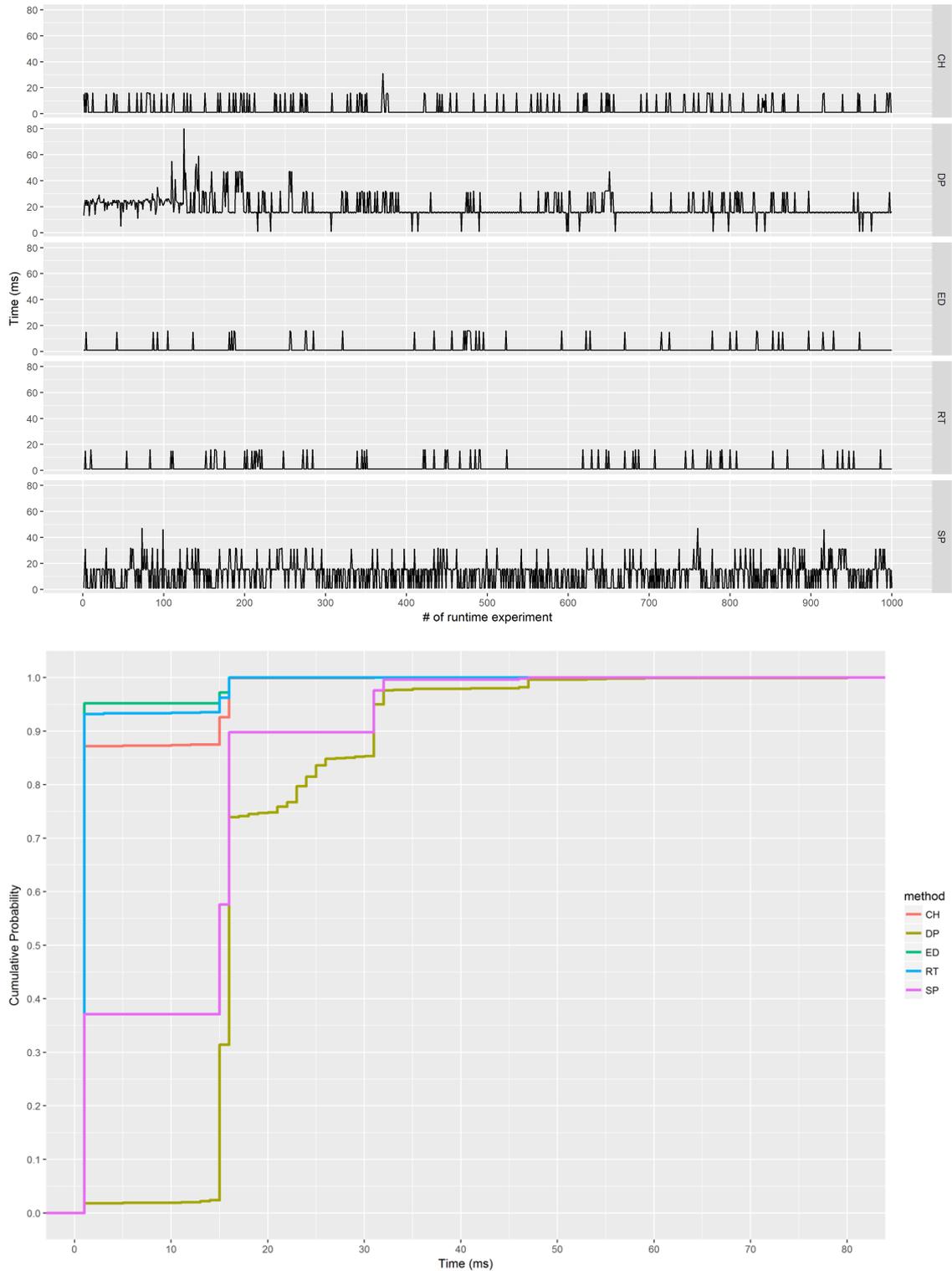

[Figure 5. The computation time comparisons among five different heuristic approaches: a) experimental runtime for each of the simulated 1000 cases; the average time differences between all approaches are statistically significant (p<0.001) using a paired two-sample t-test except for that between RT and ED; b) cumulative distribution function (CDF) of runtime.]

*5.2 Geoprocessing with traffic delay context*

In the second group of experiments, we add traffic jam information on road segments to enable context-awareness for finding a dynamic optimal meetup location under different scenarios. As shown in Figure 6a, the meetup location for two people originally is *M* (green point) as an intersection on the road network, which is derived without any traffic consideration. Figure 6b shows the new meetup location *M'* (orange point) based on the proposed geoprocessing framework with a 5-minutes traffic jam between origin $O_2$ and destination $D_2$. Figure 6c demonstrates another updated meetup location *M''* (pink point) when two people encounter a traffic jam and have total 20-minutes delay on their original shortest paths. Therefore, they should detour and meet in another optimal location to minimize the total travel time costs. Figure 6d shows an overview map of changes in the optimal meetup locations.

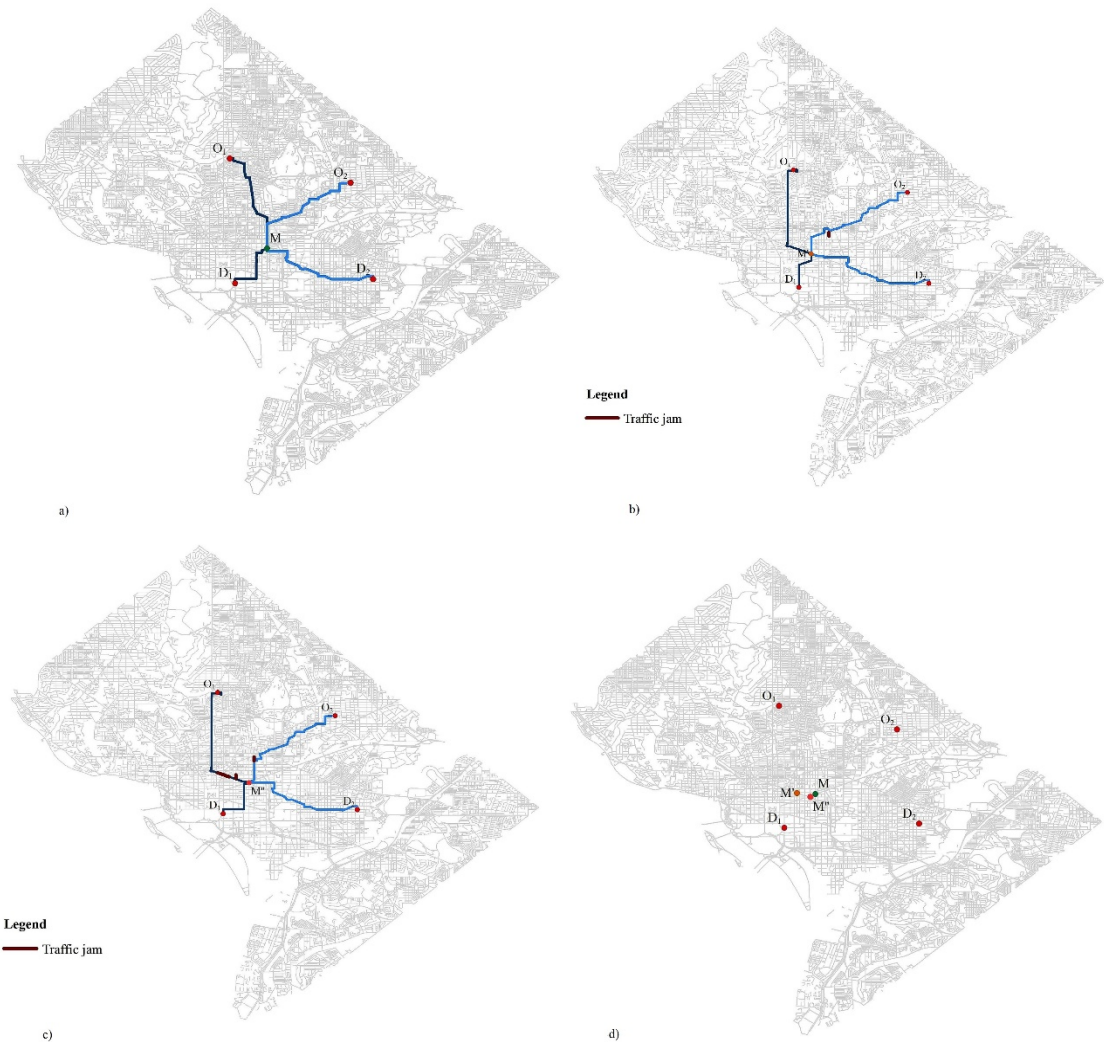

[Figure 6. The dynamic optimal meetup location for two pairs of MMO with the traffic context information: a) the meetup location M without traffic; b) the new meetup location *M'* with regard to a traffic jam from $O_2$ to $D_2$; c) the updated meetup location *M''* when both people encounter traffic jam; d) an overview of changes in the optimal meetup locations.]

One advantage of our proposed geoprocessing framework is that it can flexibly integrate external Web services or location-based data streams for adding geographic contexts. For practical engineering implementation, the access to real-time traffic alert information for a given location can help to plan optimal routes and meetup locations. Currently, there exist several public available Web services which allow developers to retrieve nearly real-time traffic information from their application programming interfaces (APIs), including *Google Maps Traffic Layer*, *Yahoo! Traffic, Esri Live Traffic, MapQuest Traffic*, and increasingly popular crowdsourcing navigation application *Waze*. However, one potential challenge is that the offline street network data or some open source street data cannot be directly matched to those online traffic services. This would require a geospatial data conflation process among various sources (Li & Goodchild 2011) or direct integration with routing service results.

## *5.3 Multi-pairs of MMO with POI information*

In the third group of experiments, to demonstrate our geoprocessing framework's capability for dealing with the case with larger than two MMO and incorporating more realistic traffic delay scenarios, we include multiple trajectories for three people with their planned origins and destinations. Note that given limited access to real-traffic jam data, we applied a traffic congestion simulation method using the OpenStreetMap road networks based on the speed performance index (see He et al. 2016). More specifically, two traffic simulation scenarios are considered: (1) Highway-based hierarchical traffic congestion simulation; and (2) All-random traffic congestion simulation. In the first scenario, we investigated historical traffic patterns in the study area using *Google Maps,* finding that most rush hour traffic jams happen along its highways. Therefore, traffic jam information along highways was included while all local roads are assigned with no traffic delay in the traffic simulation process. During the simulation, each road segment was assigned to a traffic-jam-level based on the OpenStreetMap road type [4] (i.e., *motorway = 1, trunk & primary =2, secondary =3; all other types=4*), and then the speed performance index ($R_v$) method was matched according to the Table 2 to calculate the actual passing speed ($V=V_{max}*R_v$) and the passing time under traffic for each segment. The maximum travel speed ($V_{max}$) for each road segment was assigned by utilizing the conversion rule based on the OpenStreetMap road type (in Table 3) or its attribute (*key: maxspeed*) if available. Several existing studies have also used this method in traffic assessment and routing problems when real-time traffic observation data were not available (Luxen & Vetter 2011, Geisberger et al. 2012, He et al. 2016). In the second scenario, all highway segments were randomly assigned a traffic-jam-level value (1, 2, 3, or 4) and then converted to corresponding speed performance index index ($R_v$). Such a completely random simulation may generate extremely congested traffic scenario. The street network is rendered with different colors according to the traffic level (in Figure 7), with green representing a normal speed of traffic, yellow representing moderate traffic conditions, olive indicating a slow traffic, and red indicates nearly stopped/congested traffic. During the geoprocessing, different passing speed values were added to the actual travel time calculation for the meetup location search according to the congestion level (i.e., 25% of maximum travel speed for the traffic-jam-level 1, 50% for level 2, 75% for level 3, and 100% for level 4 respectively). As shown in Figure 7a, the meetup location for three people originally is *M* (green point) at an intersection on the road network, which is derived without any traffic delay. However, Figure 7b shows a new meetup location *M'* (orange point) with regard to the hierarchical highway-based traffic jam on their ways to the original meetup location based on their shortest paths. Therefore, they should detour and meet in another optimal location to minimize total travel time cost. Moreover, Figure 7c shows an updated meetup location M'' with regard to the completely random traffic-jam context with more congested road segments, which pushed the new meetup location further away from the original one. Figure 7d provides a spatial overview of such changes in the optimal meetup locations.

---

[4] http://wiki.openstreetmap.org/wiki/Key:highway

Table 2. The criterion of traffic jam level and the speed performance index.

| Traffic-jam Level | Speed Performance Index | Traffic State | Traffic Context |
|---|---|---|---|
| 1 | [0, 0.25] | Traffic jam | The average speed is the lowest; the road traffic state is very poor. |
| 2 | (0.25, 0.50] | Slow | The average speed is low; the road traffic state is poor. |
| 3 | (0.5, 0.75] | Moderate | The average speed is moderate; the road traffic state is a little congested. |
| 4 | (0.75, 0.1] | Fast | The average speed is high; the road traffic state is good. |

Table 3. The default maximum travel speed information for OpenStreetMap road types.

| Road Type (Key: highway) | Max Speed (km/h) |
|---|---|
| motorway | 80 |
| motorway_link | 45 |
| trunk | 80 |
| trunk_link | 40 |
| primary | 65 |
| primary_link | 30 |
| secondary | 55 |
| secondary_link | 25 |
| tertiary | 40 |
| tertiary_link | 20 |
| residential | 25 |
| living_street | 10 |
| service | 15 |

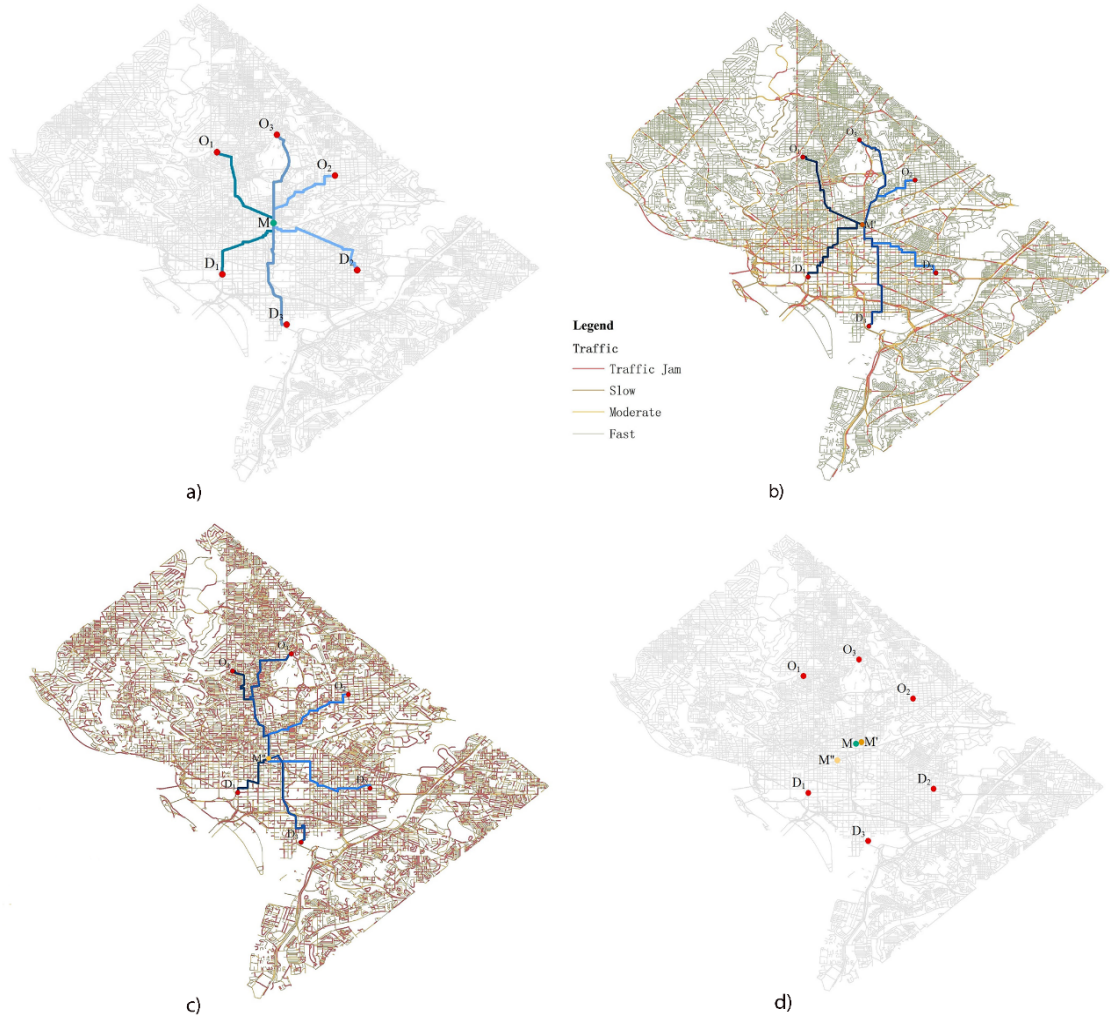

[Figure 7. The dynamic optimal meetup location for three pairs of MMO with the traffic context information: a) the meetup location M without traffic; b) the new meetup location M' with regard to hierarchical highway-based traffic jam; c) an updated meetup location M'' with regard to completely random traffic jam; d) an overview of changes in the optimal meetup locations.]

In real-world scenarios, an optimal location may be further used to integrate nearby points of interest (POI) information (e.g., parking lots, restaurants, coffee shops, and bars) for a final decision. Incorporating nearby POI with parking is necessary when the optimal meetup location is at a node that does not have parking, such as a huge roundabout. Such POI information and their spatial distributions can be retrieved using location-based services, such as *Foursquare* or *Yelp*. As shown in Figure 8, ten closest POI for each meetup location (with the name information) ranked by the distance to optimal meetup point are retrieved from the *Foursquare Venue Search API*[5]. There are 18 records in total since two venues *SiriusXM Satellite Radio* and *Hyatt Place Washington DC* are shared by two optimal meetup locations. The closest POI for *M* (about 30m away) is a bar named as *Wicked Bloom* and the closest POI for *M'* (about 145m away) is an entertainment venue *SiriusXM Satellite Radio.* The venues closer to the original meetup location *M* are represented as green markers while the venues closer to the updated meetup location *M'* are represented as orange markers. A group of people in this experiment can easily pick up one venue as the final meetup location. Moreover, since most of location-based social network applications like *Foursquare* also capture users' online check-ins and social rating behavior, which can be taken as a POI popularity score (McKenzie et al. 2015) and integrated into our geoprocessing framework to guide the geographic information observatory and spatial-social optimal decision making.

---

[5] https://developer.foursquare.com/docs/venues/search

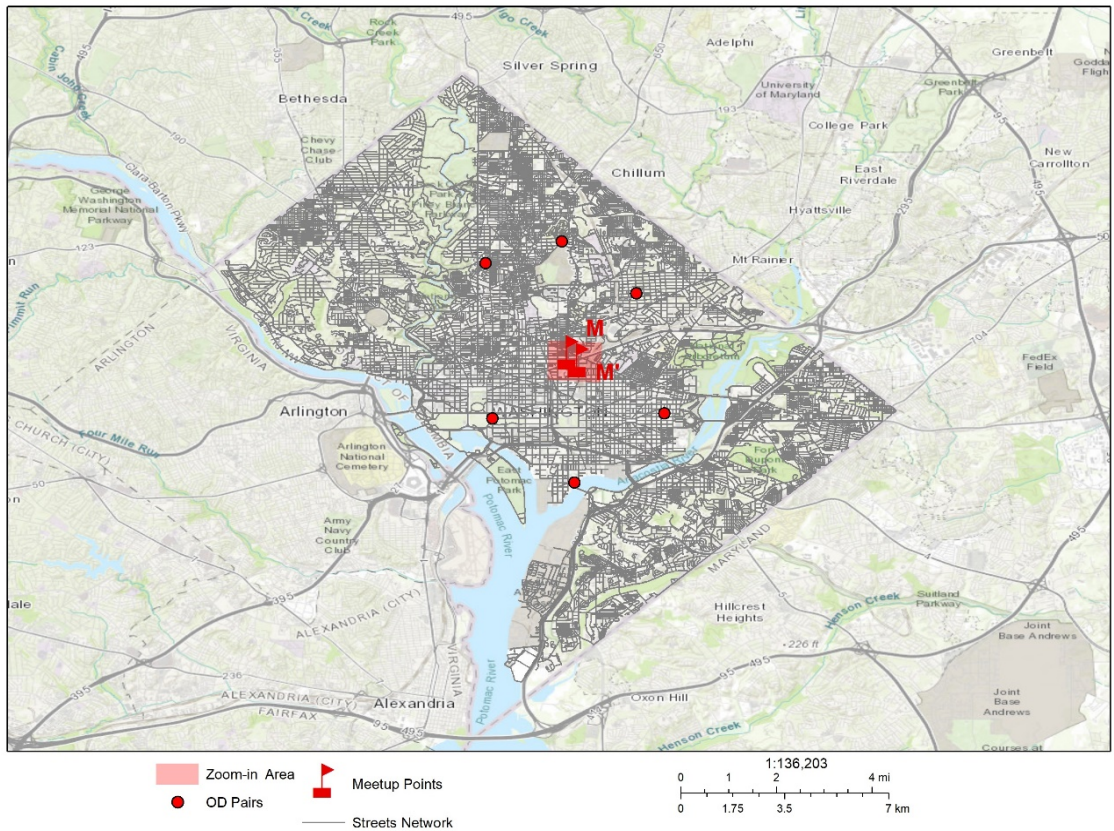

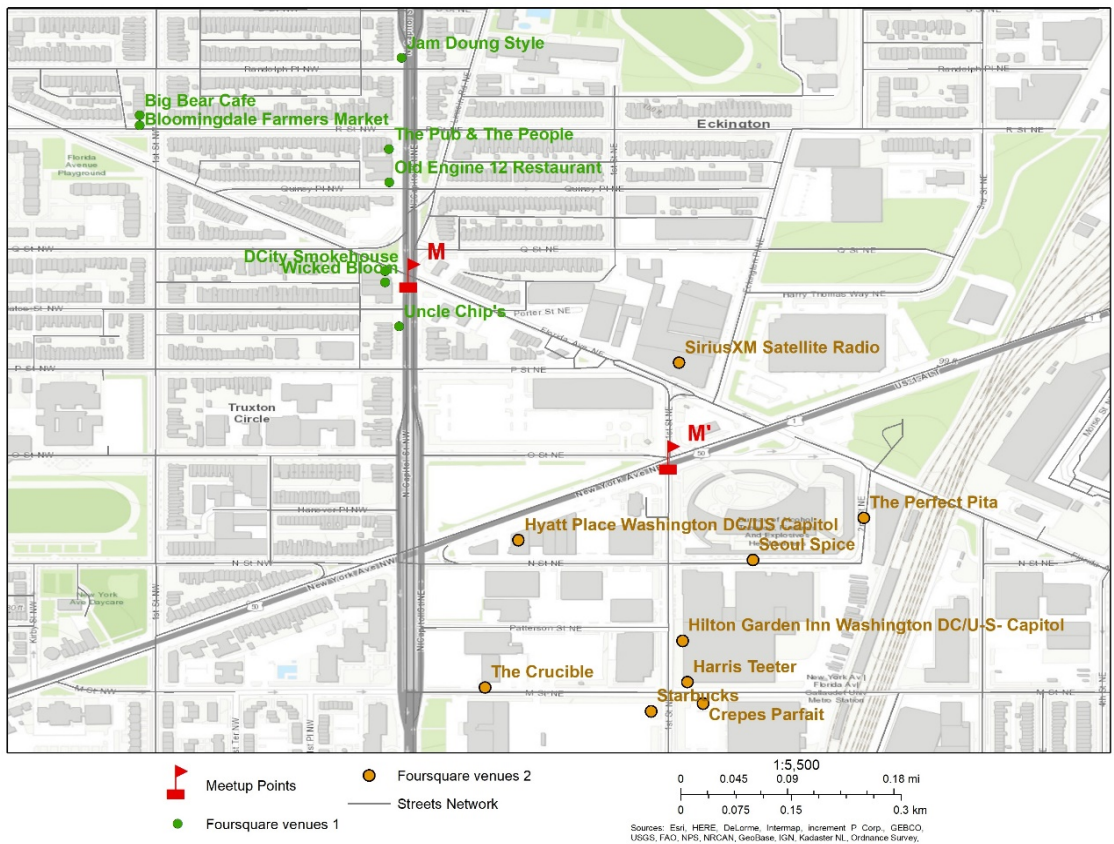

[Figure 8. The spatial distribution of meetup locations with traffic context and the nearby POI extracted from *Foursquare*.]

In this experiment, we first search for an optimal intersection on the road network by minimizing the total travel time for all MMO, and then check nearby POI information to find a venue to meet. This process may select a location which is not the actual optimal solution based on the intention of minimizing total travel costs. However, we believe that the proposed strategy has merit regarding practical POI data accessibility and API limits. Moreover, our approach allows flexible choice for a meetup POI close to the optimal point along the road network. Although such an approach does not guarantee optimality, existing studies have showed that the optimality deviation percentages at about 100~200 meters away for the p-median are approximately less than 1% or 2% (Murray 2003), which are most likely acceptable in real-world applications. If one can get full access to the whole study area POI database, it may not be necessary to run through our geoprocessing framework on road networks. Instead, the optimal meetup location (i.e., a venue) can be directly found by iterating all possible POI in the database and ranking the total travel cost for every venue. Individual preferences on certain categories of POI, the social ratings or the environment for a meetup can also be integrated into the geoprocessing framework during data cleaning, filtering and searching processes.

## 6. Conclusions and Future Work

In this research, we formalize the problem of finding the dynamic optimal meetup location for multiple moving objects into spatial optimization models using the Manhattan distance and the road network distance. We introduce a context-based geoprocessing heuristic framework to solve this problem. As a proof of concept, experiments were conducted for finding the dynamic optimal meetup location for two pairs and three pairs of MMO with different traffic delay contexts and POI information extracted from location-based social networks. Five heuristic approaches were developed and tested for the optimal location search process. The DP and RT algorithms were found to have very high accuracy for finding the optimal meetup location. However, the DP algorithm needs additional time for constructing the convex hull and tracing the diameter points for a set of MMO origins and destinations. Decision making for an efficient optimal solution in practice might be a trade-off between computation time and accuracy, especially for large-scale road networks with millions of intersections and segments. The RT algorithm runs much faster and maintains higher accuracy, so might be a better choice than other approaches (i.e., SP, CH, and ED). In addition, other improved solution techniques may be possible in future research.

This framework is implemented in a GIS environment so that it is convenient for the integration with external geographic contextual information, e.g., temporary road barriers, points of interest, and real-time traffic information when dynamically searching for optimal solutions. The proposed method can be applied in trip planning, carpooling services, and logistics management, specifically in the new era of the sharing economy.

In future work, we plan to conduct experiments on larger-sized regional road networks in other study areas in order to assess the scalability of our geoprocessing framework. In addition, the objective function in this study was specified as minimizing total travel cost for MMO; another way of defining this problem could be minimizing the average waiting time for meetup with regard to MMO's spatiotemporal constraints. This will be another interesting issue that is worth further investigation. Last but not least, we only consider one optimal meetup location in this research. Finding multiple meetup locations in a sequential order for MMO will be considered in our future research. We believe that the development of dynamic optimal meetup location search methodologies and techniques under spatiotemporal contexts can offer insights in advancing location-based services and ride-sharing applications.

**Acknowledgments**

Dr. Shaohua Wang would like to thank the funding support from the National Postdoctoral International Exchange Program of China (No. 20150081). Dr. Song Gao would like to thank the support for this research provided by the Office of the Vice Chancellor for Research and Graduate Education at the University of Wisconsin-Madison with funding (No. 135-AAC5663) from the Wisconsin Alumni Research Foundation.

**References**

Aingworth, D., Chekuri, C., Indyk, P., & Motwani, R. (1999). Fast estimation of diameter and shortest paths (without matrix multiplication). *SIAM Journal on Computing*, 28(4), 1167-1181.

Arsanjani, J. J., Zipf, A., Mooney, P., & Helbich, M. (2015). An introduction to OpenStreetMap in Geographic Information Science: Experiences, research, and applications. In *OpenStreetMap in GIScience* (pp. 1-15). Springer.

Beckmann, N., Kriegel, H. P., Schneider, R., & Seeger, B. (1990, May). The R*-tree: an efficient and robust access method for points and rectangles. In *ACM SIGMOD Record* (Vol. 19, No. 2, pp. 322-331). ACM.

Bentley, J. L. (1975). Multidimensional binary search trees used for associative searching. *Communications of the ACM*, 18(9), 509-517.

Black, J. A., Paez, A., & Suthanaya, P. A. (2002). Sustainable urban transportation: performance indicators and some analytical approaches. *Journal of urban planning and development, 128(4)*, 184-209.

Boeing, G. (2017). OSMnx: New methods for acquiring, constructing, analyzing, and visualizing complex street networks. *Computers, Environment and Urban Systems*, 65, 126-139.

Buchin, M., Dodge, S., & Speckmann, B. (2012). Context-aware similarity of trajectories. In *International Conference on Geographic Information Science* (pp. 43-56). Springer Berlin Heidelberg.

Church, R. L., & Murray, A. T. (2009). *Business site selection, location analysis, and GIS*. Hoboken, NJ: John Wiley & Sons.

Cooper, L. (1968). An extension of the generalized Weber problem. *Journal of Regional Science*, 8(2), 181-197.

Couclelis, H., Golledge, R. G., Gale, N., & Tobler, W. (1987). Exploring the anchor-point hypothesis of spatial cognition. *Journal of Environmental Psychology*, 7(2), 99-122.

De Berg, M., Van Kreveld, M., Overmars, M., & Schwarzkopf, O. C. (2000). Computational geometry. In *Computational geometry* (pp. 1-17). Springer Berlin Heidelberg.

Demšar, U., Buchin, K., Cagnacci, F., Safi, K., Speckmann, B., Van de Weghe, N., Weiskopf, D. and Weibel, R. (2015). Analysis and visualisation of movement: an interdisciplinary review. *Movement ecology*, 3(1), 5.

Dijkstra, E. W. (1959). A note on two problems in connexion with graphs. *Numerische mathematik*, 1(1), 269-271.

Dodge S, Weibel R, Ahearn SC, Buchin M, and Miller JA. (2016). Analysis of movement data, *International Journal of Geographical Information Science*, 30 (5): 825-834.

Douglas, D. H. (1994). Least-cost path in GIS using an accumulated cost surface and slopelines. *Cartographica: the international journal for Geographic Information and Geovisualization*, 31(3), 37-51.

Drezner, Z., & Goldman, A. J. (1991). On the set of optimal points to the Weber problem. *Transportation science*, 25(1), 3-8.

Geisberger, R., Sanders, P., Schultes, D., & Delling, D. (2008, May). Contraction hierarchies: Faster and simpler hierarchical routing in road networks. In *International Workshop on Experimental and Efficient Algorithms* (pp. 319-333). Springer Berlin Heidelberg.

Geisberger, R., Sanders, P., Schultes, D., & Vetter, C. (2012). Exact routing in large road networks using contraction hierarchies. *Transportation Science*, 46(3), 388-404.

Goldberg, A. V., & Harrelson, C. (2005, January). Computing the shortest path: A search meets graph theory. In *Proceedings of the sixteenth annual ACM-SIAM symposium on Discrete algorithms* (pp. 156-165). Society for Industrial and Applied Mathematics.

Hägerstraand, T. (1970). What about people in regional science? *Papers in Regional Science*, 24(1), 7-24.

Hakimi, S. L. (1965). Optimum distribution of switching centers in a communication network and some related graph theoretic problems. *Operations Research*, 13(3), 462-475.

He, F., Yan, X., Liu, Y., & Ma, L. (2016). A traffic congestion assessment method for urban road networks based on speed performance index. *Procedia Engineering*, 137, 425-433.

Holzer, M., Schulz, F., & Wagner, D. (2009). Engineering multilevel overlay graphs for shortest-path queries. *Journal of Experimental Algorithmics*, 13, 5.

Hong, I., & Murray, A. T. (2013). Efficient measurement of continuous space shortest distance around barriers. *International Journal of Geographical Information Science*, 27(12), 2302-2318.

Haklay, M. (2010). How good is volunteered geographical information? A comparative study of OpenStreetMap and Ordnance Survey datasets. *Environment and planning B: Planning and design,* 37(4), 682-703.

Hart, P. E., Nilsson, N. J., & Raphael, B. (1968). A formal basis for the heuristic determination of minimum cost paths. *IEEE transactions on Systems Science and Cybernetics*, 4(2), 100-107.

Huang, B., Wu, Q., & Zhan, F. (2007). A shortest path algorithm with novel heuristics for dynamic transportation networks. *International Journal of Geographical Information Science*, 21(6), 625-644.

Kennedy, C., Miller, E., Shalaby, A., Maclean, H., & Coleman, J. (2005). The four pillars of sustainable urban transportation. *Transport Reviews*, 25(4), 393-414.

Kim, H. M., & Kwan, M. P. (2003). Space-time accessibility measures: A geocomputational algorithm with a focus on the feasible opportunity set and possible activity duration. *Journal of Geographical Systems*, 5(1), 71-91.

Kuijpers, B., & Othman, W. (2009). Modeling uncertainty of moving objects on road networks via space–time prisms. *International Journal of Geographical Information Science*, 23(9), 1095-1117.

Kuijpers, B., Miller, H. J., Neutens, T., & Othman, W. (2010). Anchor uncertainty and space-time prisms on road networks. *International Journal of Geographical Information Science*, 24(8), 1223-1248.

Li, L., & Goodchild, M. F. (2011). An optimisation model for linear feature matching in geographical data conflation. *International Journal of Image and Data Fusion*, 2(4), 309-328.

Lombard, K., & Church, R. L. (1993). The gateway shortest path problem: generating alternative routes for a corridor location problem. *Geographical systems*, 1(1), 25-45.

Luxen, D., & Vetter, C. (2011, November). Real-time routing with OpenStreetMap data. In *Proceedings of the 19th ACM SIGSPATIAL international conference on advances in geographic information systems* (pp. 513-516). ACM.

McKenzie, G., Janowicz, K., Gao, S., Yang, J. A., & Hu, Y. (2015). POI pulse: A multi-granular, semantic signature–based information observatory for the interactive visualization of big geosocial data. Cartographica: *The International Journal for Geographic Information and Geovisualization*, 50(2), 71-85.

Miller, H. J. (1991). Modelling accessibility using space-time prism concepts within geographical information systems. *International Journal of Geographical Information System*, 5(3), 287-301.

Miller, H. J. (2005a). Necessary space—time conditions for human interaction. *Environment and Planning B: Planning and Design*, 32(3), 381-401.

Miller, H. J. (2005b). A measurement theory for time geography. *Geographical analysis*, 37(1), 17-45.

Murray, A.T. (2003). Site placement uncertainty in location analysis. *Computers, Environment and Urban Systems*, 27(2), 205-221.

Plastria, F. (1995). Continuous location problems: research, results and questions. In: Zvi Drezner, eds. *Facility location: a survey of applications and methods*, 225-262, Springer.

Samet, H., Sankaranarayanan, J., & Alborzi, H. (2008, June). Scalable network distance browsing in spatial databases. In *Proceedings of the 2008 ACM SIGMOD international conference on Management of data* (pp. 43-54). ACM.

Schwanen, T., & Kwan, M. P. (2008). The Internet, mobile phone and space-time constraints. *Geoforum*, 39(3), 1362-1377.

Shaw, S. L., & Yu, H. (2009). A GIS-based time-geographic approach of studying individual activities and interactions in a hybrid physical–virtual space. *Journal of Transport Geography*, 17(2), 141-149.

Song, Y., Miller, H. J., Zhou, X., & Stempihar, J. (2017). Green Accessibility: Estimating the Environmental Costs of Network-Time Prisms for Sustainable Transportation Planning. In *Transportation Research Board 96th Annual Meeting*, (No. 17-04713).

Siła-Nowicka, K., Vandrol, J., Oshan, T., Long, J. A., Demšar, U., & Fotheringham, A. S. (2016). Analysis of human mobility patterns from GPS trajectories and contextual information. *International Journal of Geographical Information Science*, 30(5), 881-906.

Stucky, J. (1998). On applying viewshed analysis for determining least-cost paths on digital elevation models. *International Journal of Geographical Information Science*, 12(8), 891-905.

Tellier, L. N. (1972). The Weber problem: solution and interpretation. *Geographical Analysis*, 4(3), 215-233.

Tong, D., & Murray, A. T. (2012). Spatial optimization in geography. *Annals of the Association of American Geographers*, 102(6), 1290-1309.

Vardi, Y., & Zhang, C.-H. (2001). A modified Weiszfeld algorithm for the Fermat-Weber location problem. *Mathematical Programming*, 90(3), 559-566.

Wagner, D., & Willhalm, T. (2007, February). Speed-up techniques for shortest-path computations. In *Annual Symposium on Theoretical Aspects of Computer Science* (pp. 23-36). Springer Berlin Heidelberg.

Wang, S., Zhong, E., Xiaohu, Z., Xun, Z., & Qijun, L. (2013). Shortest Path Algorithm Accelerated Technology and Search Space Analysis. *Geospatial Information*, 6, 62-65.

Wang, Y., Kutadinata, R., & Winter, S. (2016, October). Activity-based ridesharing: increasing flexibility by time geography. In Proceedings of the 24th ACM SIGSPATIAL *International Conference on Advances in Geographic Information Systems* (pp. 1-10). ACM, doi: 10.1145/2996913.2997002.

Weiszfeld, E. (1937). Sur le point pour lequel la somme des distances de n points donnés est minimum. *Tohoku Mathematical Journal*, First Series, 43, 355-386.

Xu, Z., & Jacobsen, H. A. (2010, June). Processing proximity relations in road networks. In *Proceedings of the 2010 ACM SIGMOD international conference on management of data* (pp. 243-254). ACM.

Yan, D., Zhao, Z., & Ng, W. (2011). Efficient algorithms for finding optimal meeting point on road networks. *Proceedings of the VLDB Endowment*, 4(11).

Yan, D., Zhao, Z., & Ng, W. (2015). Efficient processing of optimal meeting point queries in Euclidean space and road networks. *Knowledge and Information Systems*, 42(2), 319-351.

Zeng, W., & Church, R. L. (2009). Finding shortest paths on real road networks: the case for A*. *International Journal of Geographical Information Science*, 23(4), 531-543.

**Table List**

**Figure List**

Figure 1. A context-based geoprocessing framework to find an ideal meetup location for multiple moving objects.

Figure 2. Finding an optimal meetup location for two moving objects based on the shortest-path search space method with three scenarios: a) their shortest paths have shared segments and search spaces also overlap; b) only the shortest-path search spaces overlap; c) neither the shortest paths nor the search spaces overlap.

Figure 3. Finding an optimal meetup location for three moving objects based on five different search space strategies: a) SP; b) CH; c) DP; d) RT; e) ED.

Figure 4. The surface of the sums of road network distances for meetups located at the intersections on a road network given two fixed OD pairs.

Figure 5. The computation time comparisons among five different heuristic approaches: a) experimental runtime for each of the simulated 1000 cases; the average time differences between all approaches are statistically significant ($p<0.001$) using a paired two-sample t-test except for that between RT and ED; b) cumulative distribution function (CDF) of runtime.

Figure 6. The dynamic optimal meetup location for two pairs of MMO with the traffic context information: a) the meetup location M without traffic; b) the new meetup location M' with regard to a traffic jam from $O_2$ to $D_2$; c) the updated meetup location M'' when both people encounter traffic jam; d) an overview of changes in the optimal meetup locations.

Figure 7. The dynamic optimal meetup location for three pairs of MMO with the traffic context information: a) the meetup location M without traffic; b) the new meetup location M' with regard to hierarchical highway-based traffic jam; c) an updated meetup location M'' with regard to completely random traffic jam; d) an overview of changes in the optimal meetup locations.

Figure 8. The spatial distribution of meetup locations with traffic context and the nearby POI extracted from *Foursquare*.